\documentclass[10pt,twocolumn,letterpaper]{article}

\usepackage{cvpr}              

\usepackage[dvipsnames]{xcolor}

\definecolor{cvprblue}{rgb}{0.21,0.49,0.74}
\usepackage[pagebackref,breaklinks,colorlinks,citecolor=cvprblue]{hyperref}
\usepackage{amsmath}
\usepackage{amssymb}
\usepackage{multirow}
\usepackage{multicol}
\usepackage{booktabs}

\makeatletter
\renewcommand{\footnoterule}{%
  \kern-3\p@
  \hrule width .32 \textwidth height .4\p@
  \kern2.6\p@}
\makeatother

\newcommand{\argmax}{\mathop{\mathrm{argmax}}}

\def\paperID{27} 
\def\confName{CVPR}
\def\confYear{2024}

\title{DiffInject: Revisiting Debias via Synthetic Data Generation using \\ Diffusion-based Style Injection}

\author{Donggeun Ko$^*$\\
Aim Future\\
{\tt\small sean.ko@aimfuture.ai}
\and
Sangwoo Jo$^*$\\
Minds and Company\\
{\tt\small sangwoo.jo@mnc.ai}
\and
Dongjun Lee\\
Maum AI\\
{\tt\small akm5825@maum.ai}
\and
Namjun Park, Jaekwang Kim$^\dagger$\\
Convergence Program of Social Innovation, Sungkyunkwan University\\
{\tt\small \{951011jun, linux\}@skku.edu}
}

\begin{document}
\maketitle

\def\thefootnote{*}\footnotetext{Equal Contribution \quad  $^\dagger$Corresponding author}
\def\thefootnote{\arabic{footnote}}

\begin{abstract}
Dataset bias is a significant challenge in machine learning, where specific attributes, such as texture or color of the images are unintentionally learned resulting in detrimental performance. To address this, previous efforts have focused on debiasing models either by developing novel debiasing algorithms or by generating synthetic data to mitigate the prevalent dataset biases. However, generative approaches to date have largely relied on using bias-specific samples from the dataset, which are typically too scarce. In this work, we propose, DiffInject, a straightforward yet powerful method to augment synthetic bias-conflict samples using a pretrained diffusion model. This approach significantly advances the use of diffusion models for debiasing purposes by manipulating the latent space. Our framework does not require any explicit knowledge of the bias types or labelling, making it a fully unsupervised setting for debiasing. Our methodology demonstrates substantial result in effectively reducing dataset bias.
\end{abstract}    
\section{Introduction}
\label{sec:intro}

Deep learning networks and algorithms often inadvertently learn biases from extensive benchmark datasets. These biases, such as textures or colors, enable models to adopt shortcuts, leading to incorrect image classification. For instance, in a scenario where the majority of images depicting alligators are set against backgrounds of rivers or ponds, with scant examples of alligators on land, deep learning classifiers may rely on background features (e.g., river equates to alligator, land equates to horse) as "shortcuts" for classification. In this context, we classify the background (river or land) as \textit{task-irrelevant} features, whereas the subject of the image (the alligator) is a \textit{task-relevant} feature. Features such as skin tones, genders, and colors also constitute \textit{task-irrelevant} features that can obstruct the classifiers' ability to accurately represent objects within an image.

Prior approaches to reducing bias have explored supervised learning techniques, which depend on the annotation of biases or labels. An alternative strategy involves unsupervised learning, where generative models are employed to enhance the biased data without prior knowledge of the biases. This presents a significant advantage, as it suggests that bias mitigation should be performed without direct human supervision.

Diffusion-based models~\cite{ho2020denoising, nichol2021improved, dhariwal2021diffusion}, have demonstrated superior performance compared to GANs~\cite{karras2020stylegan2, goodfellow2014generative} in generating synthetic images. In particular, variants of Stable Diffusion~\cite{rombach2022high} have achieved remarkable success in producing high-quality synthetic images. Recent research~\cite{azizi2023synthetic, zhou2023improving, tian2023stablerep, fan2023scaling} has shown that synthetic datasets generated by these models can significantly contribute to the learning or enhancement of visual representations in deep learning models. Consequently, leveraging generative models to translate bias-conflict features and augment data presents a promising avenue for enabling biased image classifiers to accurately learn and represent bias-conflict features.

In this paper, we propose a novel framework, DiffInject, where we ``inject'' or translate bias-conflict features into the data sample and generate synthetic dataset via leveraging the diffusion model. Our framework is composed of four major steps: 1) Overfit an image classifier into the biased dataset and extract samples with top-$K$ loss in which we assume are bias-conflict images, 2) Train a diffusion model using a pronounced image benchmark dataset that generally encompasses the domain of the biased benchmark dataset, 3) Translate or inject the bias-conflict features content by leveraging the \textit{h-space} of the diffusion model and translating the feature into the original bias-aligned image, 4) Debias the biased-classifier with the augmented dataset. Injecting the content of the top-$K$ loss samples will allow the data to be translated and follow the distribution of bias-conflict images, allowing the biased model to capture the task-relevant features. 

To the best of our knowledge, we believe our method is the first to explore leveraging the diffusion model in model debiasing via unsupervised learning. Our extensive experiments demonstrate that style injection of top-$K$ loss samples to the original biased dataset allows to learn visual representation of biased-conflict samples extensively, thus debiasing the classifier.

\begin{figure*}[ht!]
  \centering
  \includegraphics[width=1.0\linewidth]{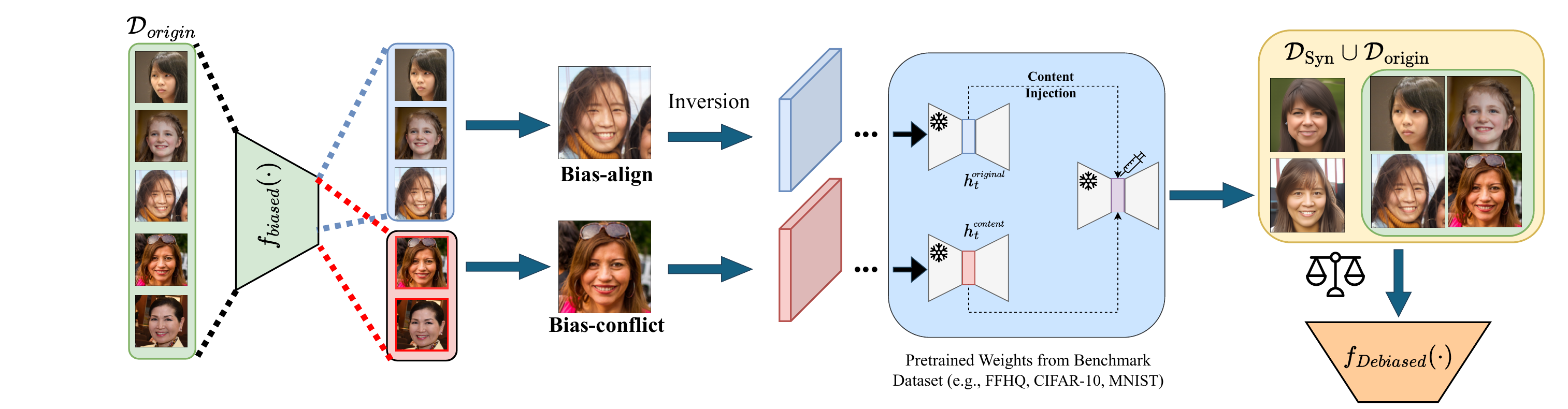}
  \caption{Overall framework of our proposed method, DiffInject.}
  \label{dataset_explanation}
\end{figure*}

\section{Method}

\subsection{Preliminaries}
\textbf{Extracting Samples with High Losses. }To extract bias-conflict samples from the dataset, we select top-$K$ loss samples, where $K$ denotes number of samples with top loss values, by intentionally overfitting the classifier where we named as bias classifier, $f_B$. By utilizing generalized cross-entropy (GCE) loss~\cite{zhang2018generalized}, we train the classifier from scratch to maximize the representation to become biased by prioritizing samples in the dataset that are ``easy-to-learn''~\cite{nam2020learning_LfF, an20222} with high probability values which we believe are \textit{bias-align} samples. Thus, GCE loss is formulated as follows:

\begin{equation}
\label{gce_loss}
    \mathcal{L}_{GCE} (p(x;\theta),y) = \frac{1-p_{y}(x;\theta)^q}{q},
\end{equation}
where $p(x;\theta),y)$ denotes softmax output of the classifier, $p_y(x;\theta)$  is the probability of the target attribute $y$ and $q \in (0,1]$ is a hyperparameter which controls the strength of amplification to make the model ``easy-to-learn''. Classifier's parameters are denoted as $\theta$. The GCE loss allows the classifier to gradually become biased by maximizing the weights on the gradients of the samples with higher probability $p_y$, formulated as follows:

\begin{equation}
    \frac{\partial GCE (p(x;\theta),y)}{\partial\theta} = p_y(x;\theta)^q \cdot \frac{\partial CE (p(x;\theta),y)}{\partial\theta}
\end{equation}

\noindent\textbf{Overfitting the Classifier to become Biased. }This allows the model have higher GCE losses and maximizes the loss when the model sees \textit{bias-conflict} samples, eventually learning shortcuts provided by abundant bias-align samples in the dataset.

After overfitting the classifier to become biased, we are able to select samples with top-$K$ losses by calculating cross-entropy loss of all the train data output. We chose $K$ as 10 following the method of AmpliBias~\cite{ko2023amplibias} and for fair comparison with other generative model methods as well. Selection process of the top-$K$ loss samples from the overfitted classifier is as follows: 
\begin{equation}
    x_{\text{c}} = \argmax_{\substack{x_i \in D_{\text{biased}}}} CE(f_B(x_i), y_i),
    \quad \text{where } x_{\text{c}} \in \mathcal{X}_{\text{c}},
\end{equation}

\begin{align}
\mathcal{X}_c = \{ & x_{c1}, x_{c2}, \ldots, x_{cK}\}, \text{where} \nonumber \\
& L(f_B(x_{c1})) > L(f_B(x_{c2})) > \ldots > L(f_B(x_{cK}))
\end{align}
where $x_{c}$, $x_{cK}$ and $\mathcal{X}_{c}$ denotes the extracted bias-conflict samples, $K$-th bias-conflict sample and a set of bias-conflict samples, respectively. $  L(f_B(x_{c1})) > L(f_B(x_{c2}))$ signifies that the loss of $x_{c1}$ is greater than $x_{c2}$, and top-$K$ losses are ordered from highest to lowest in $\mathcal{X}_c$.  

\subsection{Diffusion Model with P2 Weighting}
We train a diffusion model with Perception Prioritized (P2) weighting objective $L_{P2} = \sum_{t}\lambda^{P2}_t L_t$ from a standard benchmark dataset, where the loss function for each time step $t$ is defined as the following~\cite{choi2022perception}.
\begin{equation}
\begin{split}
    L_t &= D_{KL}(q(x_{t-1}|x_t,x_0) \parallel p_{\theta}(x_{t-1}|x_t)) \\
        &= \mathbb{E}_{x_0, \epsilon}[\lambda^{P2}_t||\epsilon-\epsilon_{\theta}(x_t,t)||^2]
\end{split}
\end{equation}
with the weighting scheme $\lambda^{P2}_t$ defined as 
\begin{equation}
    \lambda^{P2}_t = \frac{\lambda_t}{(k + \text{SNR}(t))^{\gamma}}, 
\end{equation}
where $\text{SNR}(t) = \alpha_t/(1-\alpha_t), \alpha_t = \prod_{s=1}^{t} (1-\beta_s)$ from standard diffusion model notation~\cite{{ho2020denoising, nichol2021improved, dhariwal2021diffusion}}. Both hyperparameters $\gamma$ and $k$ are set as 1, where $\gamma$ controls the degree of learning perceptually rich contents and $k$ determines the sharpness of the weighting scheme. We compare the quality of generated images with baseline diffusion models~\cite{ho2020denoising, nichol2021improved} and further validate the use of P2 weighting as our model for synthetic data generation.

\subsection{Injecting Biased Contents} \label{style_injection}
In our approach, we utilize images with high loss values to generate synthetic bias-conflict samples by integrating their biased content into randomly chosen bias-aligned samples. To achieve this, we manipulate the bottleneck layer of the U-Net architecture, denoted as \textit{h-space}~\cite{kwon2022diffusion}, as a method of content injection during the DDIM reverse process proposed in InjectFusion~\cite{jeong2024training}:
\begin{equation}
    x_{t-1} = \sqrt{\alpha_{t-1}} P_t(\epsilon_t^{\theta}(x_t | \tilde{h}_t)) + D_t(\epsilon_t^{\theta}(x_t)) + \sigma_t z_t
\end{equation}
where $P_t(\epsilon_t^{\theta}(x_t)|\tilde{h}_t)$ denotes the predicted $x_0$, $D_t(\epsilon_t^{\theta}(x_t))$ denotes the direction pointing to $x_t$, and $\tilde{h}_t$ is the modified \textit{h-space} replacing the original $h_t$. The bottleneck layer is modified using normalized spherical interpolation (Slerp) between $h_t$'s which proves to exhibit fewer artifacts when compared to either replacing or adding them:

\begin{equation}
\begin{split}
\tilde{h}_{t} = f(& h_t^{\text{original}}, h_t^{\text{content}}, \gamma) = \\
& \text{Slerp}\left(h_t^{\text{original}}, \frac{h_t^{\text{content}}}{\|h_t^{\text{content}}\|} \cdot \|h_t\|, \gamma\right)
\end{split}
\end{equation}

where $h_t^{\text{original}}$ and $h_t^{\text{content}}$ is the \textit{h-space} of the original and content image respectively, and $\gamma \in [0,1]$ denotes content injection ratio.

We apply content injection at both global and local levels. Local content injection is performed by masking the targeted area of the \textit{h-space} before applying Slerp, with the resulting interpolated $h_t$ subsequently inserted back into the original feature map. This image editing process occurs during the early stage of the generative process $[T, t_{\text{edit}}]$ followed by the stochastic noise injection during interval $[t_{\text{boost}}, 0]$. Then, we generate synthetic data samples to represent a certain proportion of the overall dataset, with the term ``bias-conflict ratio" used throughout this paper. Further details are provided in Appendix~\ref{implementation_details}.

\subsection{Training Unbiased Classifier}
Following the synthetic data generation method described in Section~\ref{style_injection}, we construct an unbiased dataset $D_{syn}$. Subsequently, we mitigate the bias in our biased classifier by training on a combined dataset comprising both synthetic and original data, denoted as $D_{total} = D_{syn} \cup D_{orig}$. This enables the model to learn more general visual representations of task-relevant features within in the dataset, thereby enhancing the debiasing of the learning process. It is important to note that labels for synthetic data is automatically assigned based on the bias-conflicted samples from which they were generated.

\section{Experiments}  \label{experiments}

\begin{table*}[!ht]
\setlength{\tabcolsep}{10pt}
\caption{Performance in accuracy (\%) for unbiased test sets across four benchmark datasets with varying ratios of bias-conflicting samples. Best performance is highlighted in bold and second-best is underlined, respectively.}
\label{tab2:class_acc_performance}
\centering
\resizebox{\linewidth}{!}{
\begin{tabular}{lccccccccc}
\toprule
& \multicolumn{4}{c}{Synthetic} & \multicolumn{4}{c}{Real-World} \\
\cmidrule(lr){2-5} \cmidrule(lr){6-9}
Methods & \multicolumn{2}{c}{Colored MNIST} & \multicolumn{2}{c}{Corrupted CIFAR-10} & \multicolumn{2}{c}{BFFHQ} & \multicolumn{2}{c}{Dogs \& Cats}\\
\cmidrule(r){2-3} \cmidrule(lr){4-5} \cmidrule(lr){6-7} \cmidrule(l){8-9}
& 1.0\% & 5.0\% & 1.0\% & 5.0\% & 1.0\% & 5.0\% & 1.0\% & 5.0\% \\
\midrule
Vanilla & 32.59 & 82.44 & 23.62 & 41.68 & 60.68 & 83.36 & 73.01 & 83.51 \\
LfF & 74.23 & 85.33 & 33.57 & 49.47 & 69.89 & 78.31 & 69.92 & 82.92 \\
DisEnt & 78.89 & 89.60 & \underline{36.49} & 51.88 & 66.00 & 80.68 & 69.49 & 82.78 \\
LfF + BE & 81.17 & 90.04 & 34.77 & \underline{52.16} & 75.08 & 85.48 & 81.52 & 88.60 \\
A\textsuperscript{2} & \underline{83.92} & \underline{91.64} & 27.54 & 37.60 & 78.98 & 86.22 & 71.15 & 83.07 \\
AmpliBias & 67.79 & 74.88 & \textbf{45.95} & \textbf{52.22} & \underline{81.80} & \underline{87.34} & \underline{73.30} & \underline{84.67} \\
DiffInject & \textbf{85.58} & \textbf{92.29} & 28.24 & 33.11 & \textbf{81.90} & \textbf{89.90} & \textbf{82.35} & \textbf{93.60} \\

\bottomrule
\end{tabular}
}
\label{quantitative_results}
\end{table*}

\textbf{Datasets}
We conduct experiments on four datasets with their matching class and bias attributes. Details are as follows: \textbf{Colored MNIST}: (Number-Color), \textbf{Corrupted CIFAR-10}: (Object-Noise), \textbf{BFFHQ}: (Age-Gender), \textbf{Dogs \& Cats}: (Animal-Fur Color).
\\
\textbf{Baselines}
We compare our method with vanilla network, LfF, DisEnt, BiasEnsemble, $A^2$, and AmpliBias as our baselines. Vanilla network is defined as multi-layer perceptron (MLP) with three hidden layers for Colored MNIST, and ResNet-18 for the remaining datasets. 
\\
\textbf{Implementation Details}
We pretrained ADM~\cite{dhariwal2021diffusion} with P2-weighting on four widely recognized computer vision benchmark datasets: MNIST, CIFAR-10, FFHQ, and AFHQ. Subsequently, we use the pretrained model weights to apply InjectFusion on our benchmark datasets: Colored MNIST (CMNIST), Corrupted CIFAR-10 (CCIFAR-10), BFFHQ, and Dogs $\&$ Cats. Further implementation details are described in Appendix~\ref{implementation_details}.

\begin{figure}[ht!]
  \centering
  \includegraphics[width=0.9\linewidth]{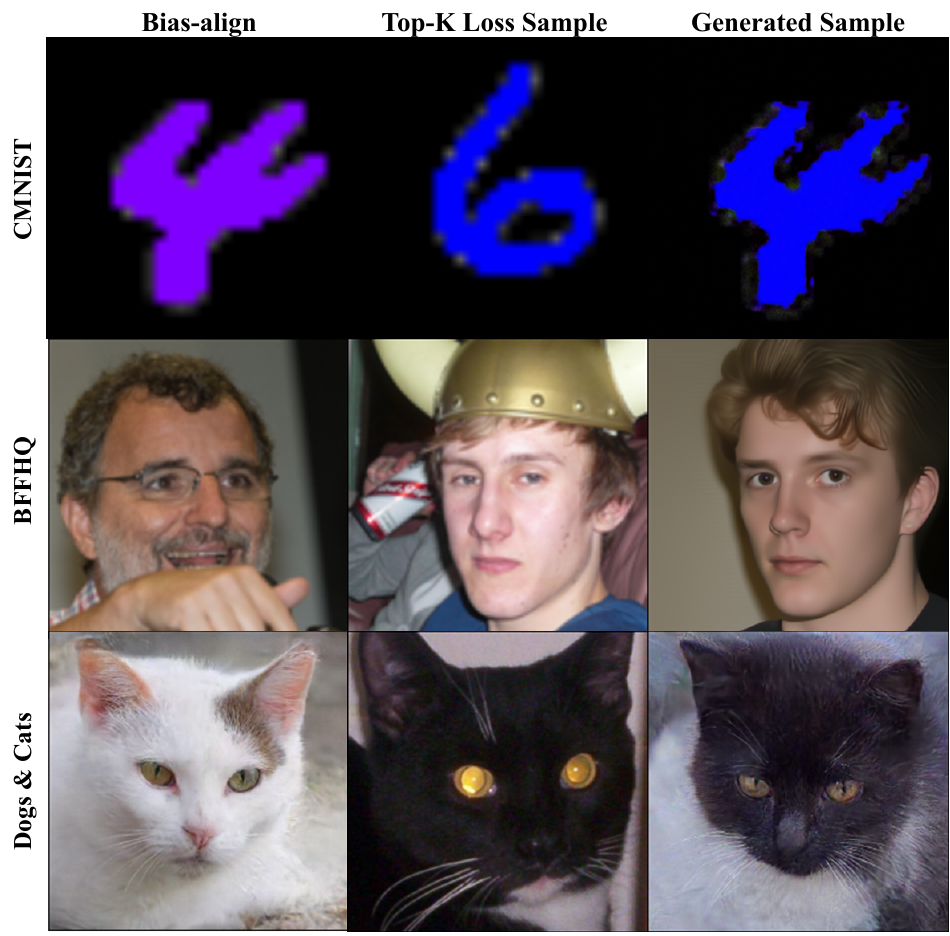}
  \caption{Generated bias-conflict samples with DiffInject. The three columns represent samples from the original dataset, top-$k$ loss samples and generated samples, respectively.}
  \label{generated_with_topK}
\end{figure}

\subsection{Analysis}
\textbf{Quantitative Results}
Table~\ref{quantitative_results} presents image classification accuracies on the unbiased test sets of synthetic datasets and real world datasets, where the ratio of bias-conflicting samples is varied at 1\% and 5\%. Most notably, DiffInject outperforms baseline methods such as LfF~\cite{nam2020learning_LfF} and DisEnt~\cite{lee2021learning_disent} on real world datasets by a substantial margin. DiffInject also achieves state-of-the-art performance on CMNIST, but lacks performance in CCIFAR-10. It is important to note that our method does not require prior knowledge in bias types and manual labeling of synthetic data, yet demonstrating superior performance across most benchmark datasets compared to the baselines. 

\begin{figure}[]
  \centering
  \includegraphics[width=0.9\linewidth]{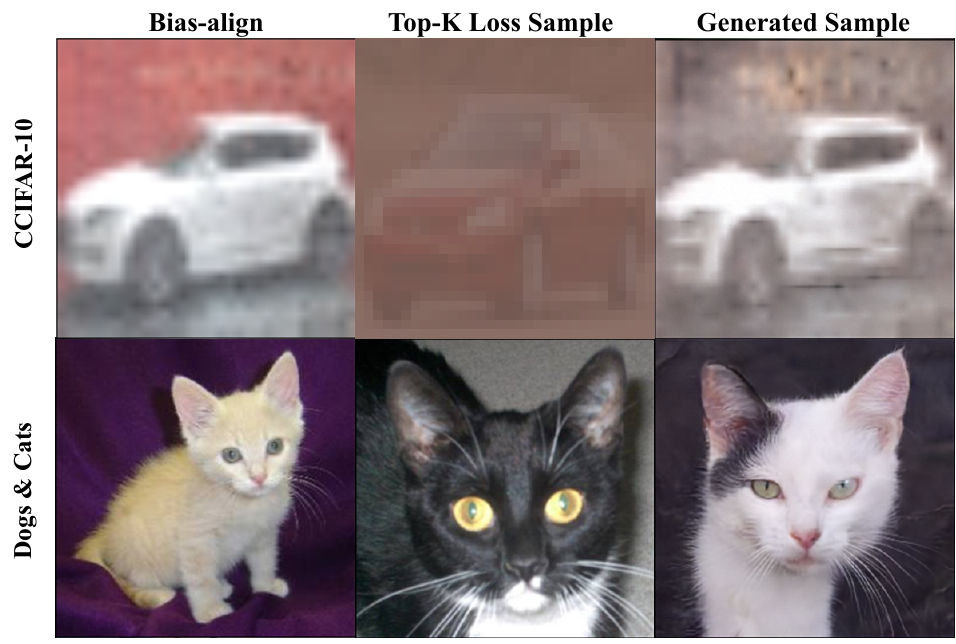}
  \caption{Generated bias-conflict samples with artifacts from our framework, DiffInject.}
  \label{generated_with_artifacts}
\end{figure}

\noindent \textbf{Qualitative Results}
We analyze the quality of synthetic samples generated from DiffInject, as illustrated in Fig.~\ref{generated_with_topK}. The three columns depict bias-aligned samples from the original dataset, original samples with top-$K$ loss, and well-generated bias-conflict samples via DiffInject, respectively. Synthetic data generated with DiffInject provide higher quality and realism, facilitating the model to learn debiased representation. For instance, the generated sample in BFFHQ effectively captures the bias-conflict attribute of ``young-male'', enriching the visual features of the dataset. Successful injection of bias-conflict attributes is also demonstrated in CMNIST and Dogs \& Cats datasets.

\noindent \textbf{Limitations} 
Figure~\ref{generated_with_artifacts} demonstrates synthetic samples generated with artifacts. To ensure fair comparison and prevent selective sampling of synthetic data, we included bias-conflict data with artifacts in $D_{syn}$ for training the classifier. For instance, in the case of Dogs \& Cats, the generated sample with artifacts fails to transfer the dark fur color observed in the bias-conflict samples. However, the presence of artifact may have positively contributed to the debiasing process, as the mixed fur color (black \& white) enables the model to learn diverse visual representation. For CCIFAR-10, our generated sample fails to effectively capture the bias attributes, which may have negatively impacted the model performance as demonstrated in Table~\ref{tab2:class_acc_performance}. Further works can be explored to successfully extract and inject texture bias attributes presented in CCIFAR-10.

\section{Conclusion}
We propose, DiffInject, a novel framework for debiasing the image classifier by augmenting synthetic data through semantic manipulation of the latent space within the diffusion model. Our approach eliminates the need for manual labeling of synthetic data and explicit knowledge of bias types in the samples, yet generating high quality bias-conflict synthetic samples. With our results demonstrating significant performance increase over benchmark datasets, we believe our work inspires the future work of leveraging diffusion models in model debiasing. 

\pagebreak

\noindent \textbf{Acknowledgement} 

\noindent This research was supported by the MSIT(Ministry of Science and ICT), Korea, under the ICAN(ICT Challenge and Advanced Network of HRD) support program(IITP-2024-RS-2023-00259497) supervised by the IITP(Institute for Information \& Communications Technology Planning \& Evaluation), Institute of Information \& communications Technology Planning \& Evaluation (IITP) grant funded by the Korea government(MSIT) (No.RS-2023-00254129, No.RS-2023-00230561, Development of Conversational Integrated AI Search Engine based on Multi-modal Technology), Graduate School of Metaverse Convergence (Sungkyunkwan University)), Aim Future, Minds and Company, and PseudoLab.

{
    \small
    \bibliographystyle{ieeenat_fullname}
    \bibliography{main}
}
\clearpage
\appendix
\setcounter{page}{1}
\maketitlesupplementary

\section{Related Work}
\textbf{Previous Debiasing Methods. }Prior approaches to debiasing have employed supervised training using explicitly defined bias labels~\cite{tartaglione2021end, sagawa2019distributionally, kim2019learning}. These methods extracted bias features and attributes from datasets under the assumption that bias labels were explicitly predefined. Recent works has sought to tackle biases without depending solely on pre-established bias labels. Instead, these strategies aim to reduce human intervention through techniques such as augmentation and re-weighting of properties. LfF~\cite{nam2020learning_LfF} pinpoints bias-conflict samples by deploying two concurrently trained and updated models, $f_D$ and $f_B$, with the debiased model $f_D$ adjusting CE loss based on a relative difficulty score. Rebias~\cite{bahng2020learning_rebias} strives to mitigate bias by disentangling and interchanging features within the latent space. A\textsuperscript{2}~\cite{an20222} leveraged StyleGAN~\cite{karras2020stylegan2} to produce augmented bias-conflict samples through a few-shot adaptation method~\cite{ojha2021few}. AmpliBias~\cite{ko2023amplibias} employed FastGAN for few-shot learning in generating synthetic bias-conflict samples. Yet, no further exploration on diffusion models were undergone in context of debiasing.

\noindent\textbf{Content injection using diffusion models} Numerous methods has been introduced to enhance controllability in image generation through diffusion-based models~\cite{ho2020denoising, nichol2021improved, dhariwal2021diffusion}. Recent works have explored the incorporation of either text guidance~\cite{ruiz2023dreambooth, gal2022image, hu2021lora, kumari2023multi, ahn2023dreamstyler} or structure guidance~\cite{zhang2023adding, mou2023t2i, li2023gligen} as a method of content injection. However, these approaches typically rely on textual descriptions or structure maps as conditioning inputs. Concurrently, alternative methodologies~\cite{kim2023reference, kwon2022diffuseit} propose the utilization of reference images for image editing guidance. In contrast, InjectFusion~\cite{jeong2024training} explores a novel approach by leveraging the latent space of a frozen, pretrained diffusion model as a means of content injection from a reference image. We further investigate the exploitation of the semantic latent space as a source of control to generate synthetic images, aiming to mitigate biases in classification tasks.

\section{Implementation Details} \label{implementation_details}
We provide further details in implementation settings as the following.

\subsection{Training ADM with P2-weighting}
We train the diffusion model by setting $T = 1000$ for all experiments. We train our model with a fixed size of 32×32 images for CMNIST and CCIFAR-10, and 256×256 for BFFHQ and Dogs $\&$ Cats. Note that images for CMNIST are resized to facilitate the implementation of P2-weighting~\cite{choi2022perception}.

\subsection{Injecting Biased Contents}
The parameter $t_{\text{edit}}$ is empirically defined such that $LPIPS(x, P_{t_{\text{edit}}}) = 0.33$, while $t_{\text{boost}}$ is fixed as 200. We set content injection ratio $\gamma$ as 0.9, 0.3, 0.7, and 0.2 for CMNIST, CCIFAR-10, BFFHQ, and Dogs $\&$ Cats, respectively. We apply local content injection for CMNIST, BFFHQ, and Dogs \& Cats, and global content injection for CCIFAR-10. Bias-conflict ratio is set as 0.6 for BFFHQ and Dogs $\&$ Cats, and 0.1 for CMNIST and CCIFAR-10. Ablation studies on bias-conflict ratio can be explored in future work.

InjectFusion~\cite{jeong2024training} takes approximately 7-10 seconds (2-3 seconds for computing inversion for each of the two images and applying content injection, respectively) per generated sample for BFFHQ and Dogs \& Cats, and 90 seconds (30 seconds for computing inversion for each of the two images and applying content injection, respectively) per generated sample for CMNIST and CCIFAR-10, based on NVIDIA A100 and NVIDIA H100 GPUs. We use multiprocessing to accelerate the content injection process. 

\subsection{Training Unbiased Classifier}
We implement the preprocessing techniques described in DisEnt~\cite{lee2021learning_disent}: We apply random crop and horizontal flip transformations for CCIFAR-10 and BFFHQ, and apply normalization with the mean of (0.4914, 0.4822, 0.4465) and standard deviation of (0.2023, 0.1994, 0.2010) for each channel. We do not implement any augmentations for CMNIST and Dogs $\&$ Cats. We use cross entropy loss as our loss function, and use Adam optimizer with the learning rate of 0.001 for CMNIST and CCIFAR-10, and 0.0001 for BFFHQ and Dogs $\&$ Cats.

\subsection{Additional Generated Synthetic Images}
\begin{figure*}[ht!] 
  \centering
  \includegraphics{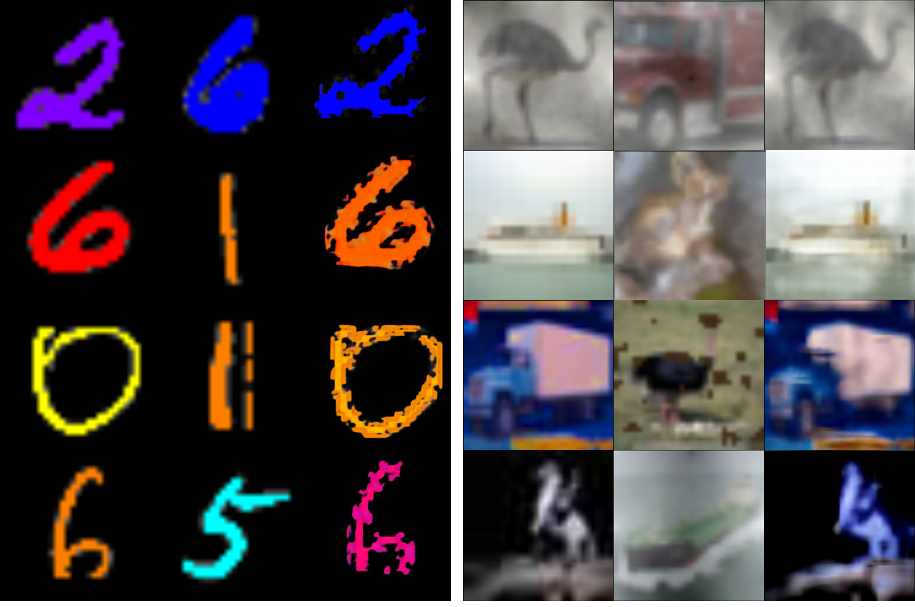}
  \caption{Examples of generated bias-conflict samples with DiffInject for CMNIST and CCIFAR-10 dataset. The three columns represent samples from the original dataset, top-$k$ loss samples and generated samples, respectively.}
  \label{appendix1_fig}
\end{figure*}

\begin{figure*}[]
  \centering
  \includegraphics{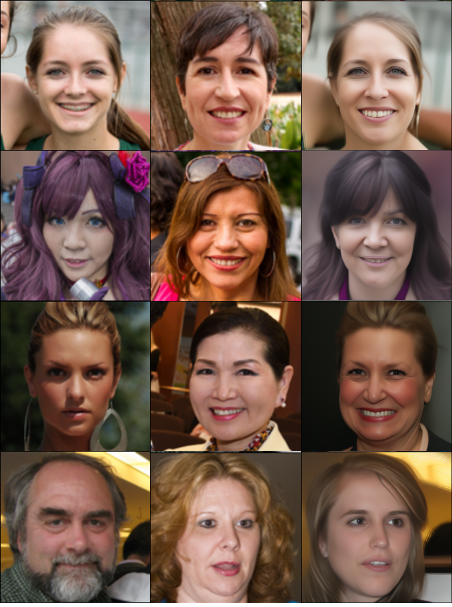}
  \caption{Examples of generated bias-conflict samples with DiffInject for BFFHQ dataset. The three columns represent samples from the original dataset, top-$k$ loss samples and generated samples, respectively.}
  \label{appendix2_fig}
\end{figure*}

\begin{figure*}[ht!] 
  \centering
  \includegraphics{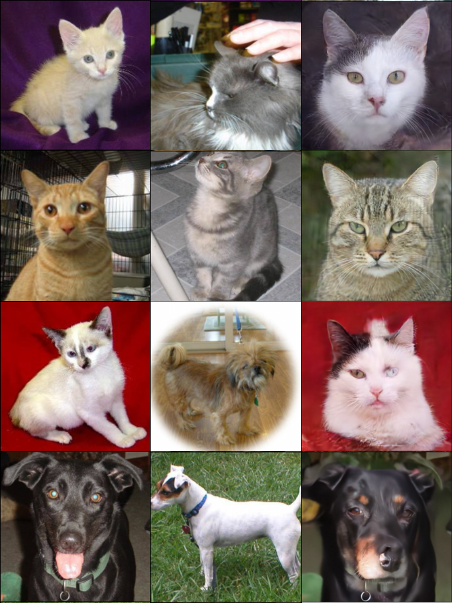}
  \caption{Examples of generated bias-conflict samples with DiffInject for Dogs \& Cats dataset. The three columns represent samples from the original dataset, top-$k$ loss samples and generated samples, respectively.}
  \label{appendix3_fig}
\end{figure*}

In this section, we provide additional samples generated from DiffInject. Figure~\ref{appendix1_fig} includes synthetic samples for CMNIST and CCIFAR-10. Figure~\ref{appendix2_fig} and Figure~\ref{appendix3_fig} includes generated samples for BFFHQ and Dogs \& Cats, respectively. Each figure consists of three columns representing, from left to right, the original samples from the dataset, top-$k$ loss samples, and synthetic samples generated from DiffInject, respectively.


\newpage

\newpage

\end{document}